\documentclass{llncs}  

\usepackage{graphicx} 
\usepackage[T1]{fontenc}
\usepackage{amsmath}
\usepackage[english]{babel}
\usepackage{array,tabularx}
\usepackage{booktabs} 
\usepackage{makeidx} 
\usepackage{xparse}
\usepackage{moredefs}
\usepackage{subfig}
\usepackage{lips}
\usepackage{epstopdf} 
\usepackage{color}
\usepackage{csquotes}
\usepackage{xcolor}
\usepackage{times}
\usepackage[hidelinks]{hyperref}
\usepackage[nolist,nohyperlinks]{acronym}
\usepackage{comment}
\usepackage{paralist}
\usepackage{cleveref}
\usepackage{wrapfig}
\usepackage{multirow}
\usepackage{microtype}
\usepackage{listings}
\usepackage[super]{nth}
\usepackage{amsfonts}

\usepackage{float}
\floatstyle{plaintop}
\restylefloat{table}

\usepackage[style=base,labelfont=bf,labelsep=period,tableposition=top,font=small]{caption}
\makeatletter
\renewcommand\@biblabel[1]{#1.}
\makeatother
\addto\captionsenglish{}

\usepackage{fancyhdr}
\fancypagestyle{WI_footer}{\fancyhf{}\fancyfoot[L]{\color{black}\scriptsize 17th International Conference on Wirtschaftsinformatik\\February 2022, Nürnberg, Germany}}

\usepackage{cite}

\crefformat{footnote}{#2\footnotemark[#1]#3}
\usepackage{xspace}
	\newcommand\ie{i.\,e.\xspace}
	\newcommand\eg{e.\,g.\xspace}

	\newcommand\cf{cf.\xspace}

	\newcommand{\mathup}[1]{\mathrm{#1}}

	\newcommand{\e}[1]{\mathup{e}^{#1}}

  \newcommand{\E}{\mathbb{E}}

\expandafter\def\expandafter\UrlBreaks\expandafter{\UrlBreaks
  \do\a\do\b\do\c\do\d\do\e\do\f\do\g\do\h\do\i\do\j%
  \do\k\do\l\do\m\do\n\do\o\do\p\do\q\do\r\do\s\do\t%
  \do\u\do\v\do\w\do\x\do\y\do\z\do\A\do\B\do\C\do\D%
  \do\E\do\F\do\G\do\H\do\I\do\J\do\K\do\L\do\M\do\N%
  \do\O\do\P\do\Q\do\R\do\S\do\T\do\U\do\V\do\W\do\X%
  \do\Y\do\Z}

\newcolumntype{L}[1]{>{\raggedright\arraybackslash}p{#1}}   
\newcolumntype{C}[1]{>{\centering\arraybackslash}p{#1}}     
\newcolumntype{R}[1]{>{\raggedleft\arraybackslash}p{#1}}    

\begin{document}
\frontmatter         
\mainmatter           

\title{Instance Selection Mechanisms for Human-in-the-Loop Systems in Few-Shot Learning}

\author{Johannes Jakubik \and
Benedikt Blumenstiel \and
Michael Vössing \and
Patrick Hemmer}

\institute{Karlsruhe Institute of Technology, Karlsruhe, Germany\\
\email{\{johannes.jakubik, benedikt.blumenstiel, michael.voessing, patrick.hemmer\}@kit.edu}
}

\maketitle
\setcounter{footnote}{0}

\begin{abstract}
Business analytics and machine learning have become essential success factors for various industries---with the downside of cost-intensive gathering and labeling of data. Few-shot learning addresses this challenge and reduces data gathering and labeling costs by learning novel classes with very few labeled data. In this paper, we design a human-in-the-loop (HITL) system for few-shot learning and analyze an extensive range of mechanisms that can be used to acquire human expert knowledge for instances that have an uncertain prediction outcome. We show that the acquisition of human expert knowledge significantly accelerates the few-shot model performance given a negligible labeling effort. We validate our findings in various experiments on a benchmark dataset in computer vision and real-world datasets. We further demonstrate the cost-effectiveness of HITL systems for few-shot learning. Overall, our work aims at supporting researchers and practitioners in effectively adapting machine learning models to novel classes at reduced costs. 
\newline

{\textbf{Keywords:} Instance selection mechanisms, Human-in-the-Loop, Few-shot learning, Computer vision}
\end{abstract}

\thispagestyle{WI_footer}


\section{Introduction}
\label{sec:introduction}

State-of-the-art supervised machine learning (ML) models generally require vast amounts of data. While model performance usually increases with more data, the gathering and labeling of data is a time-consuming endeavor that often requires the knowledge of domain experts resulting in a cost-intensive process. This observation especially applies to computer vision \cite{patterson2017deep}. Few-shot learning addresses this challenge by allowing ML models to adapt to novel classes while only requiring a small number of labeled instances for each new class \cite{Snell2017}. Despite the desirable performance of few-shot learning given the small amount of labeled data, the performance is often insufficient for real-world applications and, in addition, traditional supervised learning is not an option with very small amounts of data. Thus, given a range of incorrectly classified instances in few-shot settings, the questions arises in which order the instances should be presented to the human expert to acquire expert knowledge efficiently.

In this paper, we address this by developing a human-in-the-loop (HITL) system that utilizes state-of-the-art few-shot learning techniques. We demonstrate that superior performance and significant cost savings can be achieved by acquiring human expert knowledge through so-called instance selection mechanisms. By conducting a comprehensive evaluation of a wide range of instance selection mechanisms, we add to the ongoing discourse on collaboration between humans and AI \cite{Dellermann2019}, the increasing usage of human-in-the-loop in real-world ML and data science \cite{Dragut2021}, and the current state-of-the-art in few-shot learning. 

In the last few years, few-shot learning has been used in various fields, such as robotics (\eg, imitating users from few demonstrations), acoustic signal processing (\eg, voice conversions), computer vision (\eg, image classification, part labeling, shape view reconstruction for 3D objects), and natural language processing (\eg, translation and sentence completion) \cite{Wang2020Generalizing}. In the following, we shed light on characteristics of practical use-cases for few-shot image classification. Practitioners should consider few-shot learning in settings with low data availability, high labeling costs, when the time for model training is scarce, or when the environment requires frequent adaptions to new tasks consisting of novel classes. Examples for few-shot image classification range from labeling assistance (\cf \cite{choi2018structured}), computer vision-assisted checkouts in foodservice \cite{Wang2020Meta}, manufacturing \cite{Li2021} and production lines given varying products, stocktaking in logistics \cite{Wang2020Meta}, agriculture \cite{karami2020automatic,liu2020few,argueso2020few} or censuses of populations in biology \cite{sun2020few}. {In all of these settings, a small amount of labeled instances is provided for each novel class in order to adapt the model to these classes. For instance, when adapting checkout systems in foodservice to novel classes (\ie, new products), operators of the model take a several images for each novel class and provide labels for the classes.} Overall, our work aims at domain-independently enhancing the performance of few-shot learning in image classification by acquiring human expert knowledge within a HITL system.

Our contribution in this paper is three-fold: (1)~We extend the existing literature by providing a comprehensive evaluation of a wide range of existing and novel instance selection mechanisms for few-shot systems. Further, to the best of our knowledge, we are the first to adapt ensemble-based instance selection mechanisms to the few-shot learning setting. (2)~Our evaluation is tailored to real-world industry datasets to analyze the impact of instance selection mechanisms in real-world settings and inform practitioners about optimal instance selection strategies for efficient model adaptation. (3)~We assess the cost-effectiveness of instance selection mechanisms and, thus, account for business considerations when deploying ML models. Overall, our work thereby supports researchers and practitioners in effectively adapting ML models to novel classes with reduced labeling effort. 

The paper is structured as follows: We provide an overview of the relevant literature in Section \ref{sec:theoretical_background}. In Section \ref{sec:methodology}, we introduce our methodology. In Section \ref{sec:experimental-setup}, we present the experimental setup. We report the results of different instance selection mechanisms in different settings in Section \ref{sec:results}. Section \ref{sec:discussion} discusses implications for researchers and practitioners and further general implications for the collaboration of human experts and ML models. Finally, we suggest directions for future research.

\section{Related Work}
\label{sec:theoretical_background}

Our work is embedded in research on hybrid intelligence. Hybrid intelligence refers to the combination of the intuitive intelligence of humans and the analytical intelligence of AI systems \cite{Dellermann2019}. Research in the field of hybrid intelligence distinguishes three approaches: First, systems where AIs consult human experts to acquire knowledge about instances with an uncertain prediction outcome. Second, systems where humans leverage AIs as decision support systems. And third, systems where AI and humans perform tasks collectively \cite{Dellermann2019}. We refer to the first approach as a human-in-the-loop system, which allows AI models to request support from human experts for instances with an uncertain prediction outcome. Thereby, HITL systems try to minimize the shortcomings of AI systems and make them more effective \cite{wu2021survey,Dellermann2019}. Recently, researchers identified the strategic importance of HITL systems in organizations to ensure that the deployed ML model achieves the required performance \cite{gronsund2020augmenting}. Moreover, HITL systems gain further traction in the field of data science \cite{Dragut2021}. In addition, recent research found HITL systems to be of particular value when the availability of data is limited. This particularly applies when pre-trained models need to learn new concepts and adapt to novel tasks, as, for instance, in few-shot learning \cite{Dellermann2019}, which makes HITL and few-shot learning an ideal match. In computer vision, hybrid intelligence plays an increasingly important role. Recently, \cite{Zschech2021} outlined four mechanisms as a basis for socio-technical research on hybrid intelligence systems: Automation, signaling, manipulation, and collaboration. Our approach of HITL few-shot learning employs these mechanisms and specifically studies instance selection mechanisms as part of the collaboration and signaling mechanism. Further, the few-shot learning model refers to the automation mechanism, while the manipulation mechanism is covered by our HITL system.

From a technical perspective, only a few works examined the selection of instances in few-shot learning. Recent literature considers the selection of the initial instance per class in few-shot learning a clustering problem. In this context, Boney and Ilin \cite{Boney2017} study cluster-based instance selection mechanisms, finding strong results over the random baseline for the choice of the initial unlabeled instance per class. In contrast, our approach extends the selection of instances and, therefore, we iteratively examine the effect of acquiring human expert knowledge for the entire set of unlabeled data. Further, we provide comparisons for different levels of human labeling budget and to confidence-based and ensemble-based approaches. Pezeshkpour et al. \cite{Pezeshkpour2020} study active instance selection in the context of few-shot learning by proposing upper bounds and focusing on the comparison of batch-based vs. non-batch-based approaches. Besides a reduced number of instance selection mechanisms and less complex benchmark datasets, this work does not include real-world industry datasets or cost considerations. Additional studies focus on HITL few-shot learning in the context of reinforcement learning, and the detection of out-of-distribution instances \cite{wan2020human}. In contrast to recent works, we bring various different instance selection mechanisms together and provide a clear path to effective instance selection based on the human labeling budget and, therefore, particularly take costs into account. Further, in contrast to recent related literature, we evaluate all experiments on real-world industry datasets to inform practitioners on meaningful instance selection mechanisms.

Related to HITL systems is the Active Learning (AL) paradigm. It pursues the idea of selecting those data instances for human annotation during model training that contribute the most to its  learning process while maintaining its predictive performance \cite{settles2009active}. The most common instance selection strategies belong to the category of \textit{pool-based} approaches which can be further categorized into \textit{diversity-}, \eg, \cite{sener2017active}), \textit{uncertainty-based}, \eg, \cite{beluch2018power,gal2017deep,hemmer2020,wang2016cost}, as well as a \textit{combination} of both, \eg, \cite{ash2019deep,kirsch2019batchbald}. We want to highlight that the AL paradigm differs from our setting in the way that we do not utilize the instances selected by an expert for model training.    

In contrast to active learning, our more general approach focuses on effectively correcting the predictions of few-shot learning models based on human expert knowledge. For this, we determine the instances with the most uncertain prediction outcome through so called instance selection mechanisms, as instances with high prediction uncertainty are more likely to be misclassified.

\section{Methodology}\label{sec:methodology}

Our work uses Prototypical Neural Networks (PNN), a popular few-shot learning implementation, for image classification as proposed in \cite{Snell2017}. {The overall goal of PNNs is to adapt to novel classes based on few labeled data per each novel class. To accomplish this, PNNs aim at classifying unlabeled data from novel classes based on the closest distance to the few labeled data.}
More specifically, the processing of images in PNN is conducted in three stages: First, PNN utilize a feature extractor to compute a low dimensional representation of the input images (\ie, feature embeddings). Second, the model leverages a small set of labeled data for each {novel} class (\ie, the \textit{support} set $\mathcal{S}$) to compute a class prototype $\mathcal{P}_c$ for {novel} class $c$ defined as the mean of its respective support set. Third, the model then classifies each unlabeled data instance of the \textit{query} set $\mathcal{Q}$ by determining its nearest class prototype. Therefore, the euclidean distances are calculated between the feature embeddings of each query instance and each class prototype. Based on this, each query instance is assigned to the {novel} class with the nearest class prototype. We provide the following naming conventions: The task of the prototypical network is to classify unlabeled instances of $N$ unseen classes based on a small amount of $k$ labeled instances per class. We refer to this setting as an $N$-way $k$-shot classification task. We later refer to the size of the query by $b$-query.

\begin{figure}
    \centering
    \includegraphics[width = 0.9\linewidth]{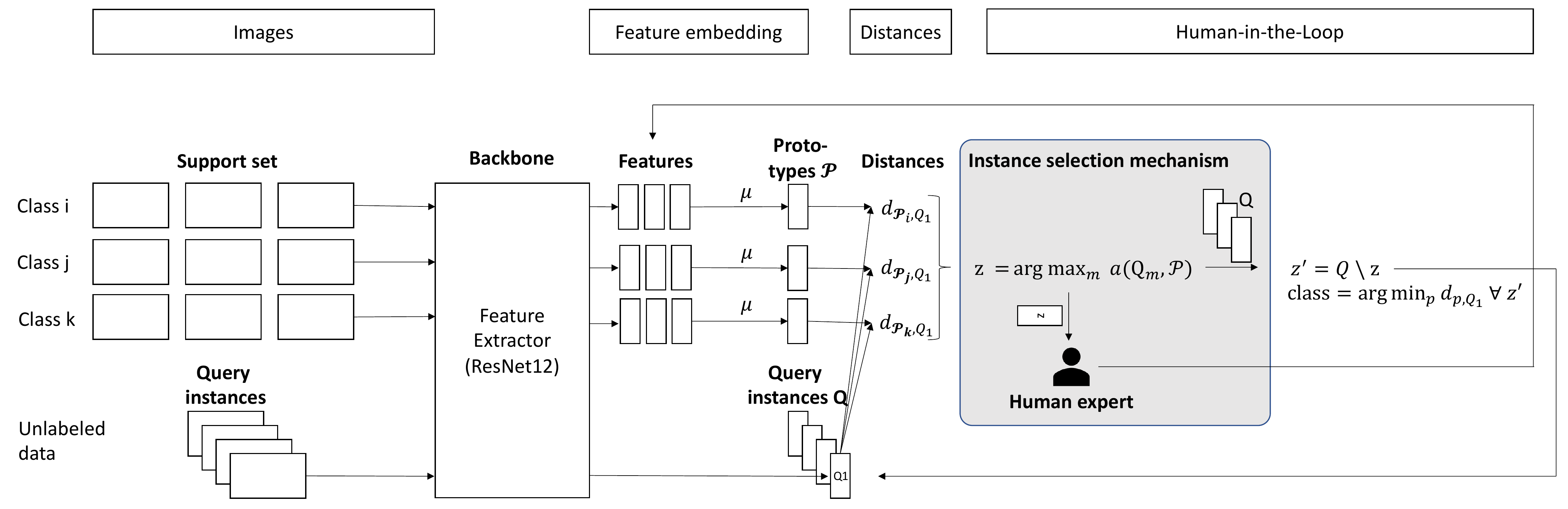}
    \caption{Proposed HITL system for few-shot image classification. Per iteration, a specific instance $z$ is selected given by the maximum model uncertainty. $\mathcal{P}_c$ represents an aggregated representative for each class, the so-called class prototype, and $\mathcal{Q}$ denotes the query set.}
    \label{fig:approach}
\end{figure}

In this work, we build a HITL system on top of the PNN. This allows us to acquire human expert knowledge when the model is uncertain about a specific instance. The procedure in the HITL system is as follows: First, an instance selection mechanism determines the instance $z$ with the highest model uncertainty. This instance is then passed to the human expert for labeling. Note that we simulate the human in our experiments and assume that the human expert labels the instance correctly. To include the obtained knowledge in the model, we add the labeled instance to the support set and re-calculate the prototype. Like this, the prototype is adjusted towards the newly labeled instance, which influences the distances of subsequent unlabeled data $z' = Q \text{\textbackslash} z$ to this prototype. Therefore, our HITL system based on PNN supports updating the best predictor at each step based on new data, which is not the case for HITL systems based on traditional classifiers. This update mechanism makes our approach a valuable technique for acquiring human expert knowledge. We summarize the structure of the HITL few-shot learning system in Figure \ref{fig:approach}. Overall, the core of the HITL system is the meaningful selection of instances to be labeled by the human expert. 

The instance selection mechanism encompasses three major steps: (1)~the selection of the most uncertain instance, (2)~acquiring human expert knowledge for this instance, and (3)~including the acquired knowledge in the prototype to enhance the classification performance on upcoming instances. The selection of the most uncertain instance in step~(1) is based on the \textit{acquisition function} $a(\cdot)$ that measures the model uncertainty on an instance level. For each query, we select instance $z$ to be labeled by the human expert by maximizing an acquisition function $z = \arg\max_m a(Q_m, \mathcal{P}_c)$ \cite{Boney2017}. In this study, we assess the performance of instance selection mechanisms based on various acquisition functions. {Specifically, we focus on single instance mechanisms (\ie, in every time step, human expert knowledge is acquired for one specific instance) and exclude batch-based mechanisms (\ie, in every time step, human expert knowledge is acquired for multiple instances) for reasons of comparability.} From the single instance mechanisms, we evaluate confidence-based, cluster-based, and ensemble-based mechanisms as well as random instance selection and a single instance oracle (see \cite{Pezeshkpour2020,Boney2017}). {We further evaluated distance-based mechanisms and Gaussian Process-based mechanisms in pre-studies and excluded them from the main analysis due to very limited performances. Overall, we aim at providing a holistic evaluation of single instance mechanisms.} In the case of confidence-based instance selection mechanisms, we first calculate the distances between the unlabeled data and the class prototypes and, second, compute the softmax score on the distances as the basis for the confidence measure following \cite{Pezeshkpour2020}. The evaluated confidence-based mechanisms are Minimum Confidence, Maximum Entropy and Margin, which are commonly used in few-shot learning and AL \cite{Pezeshkpour2020, wang2016cost}. For cluster-based instance selection mechanisms, we implement a k-Means clustering to categorize the unlabeled data as proposed in \cite{Boney2017}. {Here, we implement the selection mechanisms as proposed in \cite{Boney2017} but instead of selecting the most certain instance as in \cite{Boney2017}, we select the most uncertain instance to acquire human expert knowledge.} For the single instance oracle, we proceed as follows: At each step, the oracle separately updates the prototype for each instance from the query set and calculates the performance increase. The oracle selects then (ex-post) the instance that would have achieved the largest performance gain in this step (compare \cite{Pezeshkpour2020}). 

In addition to existing instance selection mechanisms, we propose ensemble-based instance selection mechanisms for few-shot learning. Therefore, several feature embeddings are calculated from different ensembles $e$ using Monte-Carlo Dropout \cite{gal2017deep}. For each ensemble, the class prototypes are calculated separately. The resulting prediction scores are averaged over all ensembles, while the spread of the model predictions for each instance implicitly indicate the uncertainty of the ensemble. We later specifically calculate the uncertainty of the ensemble-based approach with Variation Ratio and Bayesian Active Learning by Disagreement (BALD) {which are the two best performing ensemble-based mechanisms in \cite{gal2017deep}}. We outline all mechanisms and calculations in Table \ref{tab:overview}.

\begin{table}[ht]
    \centering
    \scriptsize
    \begin{tabular}{lp{8.7cm}}\toprule
         Instance selection & Description \\\midrule
         \textbf{Confidence-based:} & \\
         ~~- Minimum Confidence & Select instance by minimizing the highest class probability (\ie, model confidence) \\
         &  \quad $a(Q_m, \mathcal{P}_c) = - max_c\mathbb{P}(y=c|Q_m)$\\
         ~~- Maximum Entropy & Select instance by maximizing the entropy of class distributions for classifier output \\
         &   \quad $a(Q_m, \mathcal{P}_c) = \mathbb{H}[y|Q_m, \mathcal{P}_{c}] = - \sum_c \mathbb{P}(y=c|Q_m)log\mathbb{P}(y=c|Q_m)$\\
         ~~- Margin & Select instance by minimizing the difference of first and second highest class probability \\
         &  \quad $a(Q_m, \mathcal{P}_c) = - (\mathbb{P}(y=c_1(Q_m)|Q_m) - \mathbb{P}(y=c_2(Q_m)|Q_m))$ \\
         \textbf{Cluster-based:} & \\
         ~~- Cluster Maximum Distance & Select instance furthest away from the cluster center $h_{c'}$ for clusters $c'$ \\
         &  \quad $a(Q_m, c') = d(Q_m, h_{c'})$\\
         ~~- Cluster Maximum Entropy & Select instance by maximizing the entropy of cluster distributions for clusters $c'$ \\
         &  \quad $a(Q_m, c') = \mathbb{H}[y|Q_m, {c'}]$ \\
         ~~- Cluster Margin & Select instance by minimizing the difference of first and second most likely cluster $c_1, c_2$ \\
         & \quad $a(Q_m, c') = - (\mathbb{P}(y=c_1(Q_m)|Q_m) - \mathbb{P}(y=c_2(Q_m)|Q_m))$ \\ 
         & \\
         \multicolumn{2}{l}{\textbf{Ensemble-based (Proposed in this work):}} \\
         ~~- BALD & Select instance {with confident individual models but uncertain ensemble} \cite{houlsby2011BALD} \\
         & \quad $a(Q_m, \mathcal{P}_{c, e}) = \mathbb{H}[y|Q_m, \mathcal{P}_{c, e}] - E_{p(\omega)}[\mathbb{H}[y|Q_m, \omega]]$ with model weights $\omega$\\
         ~~- Variation ratio & Select the instance with highest ratio of ensemble predictions not being the mode class\\
         & \quad $a(Q_m, \mathcal{P}_{c, e}) = 1 - max_y \mathbb{P}(y|Q_m) $ \\
         & \\
        \multicolumn{2}{l}{\textbf{Baseline and upper bound:}} \\
         ~~- Oracle & Select instance that accelerates overall model accuracy most significantly \\
         ~~- Random selection & Select a random instance from the query\\\bottomrule
    \end{tabular}
    \caption{Instance selection mechanisms for HITL few-shot learning evaluated in this study.}
    \label{tab:overview}
\end{table}

\section{Experimental setup}\label{sec:experimental-setup}

In the following, we describe our experimental setup. We introduce the used datasets and the evaluation metrics, and provide details on our implementation.

\subsection{Data}
We evaluate our approach on multiple datasets. Following \cite{Snell2017}, we evaluate all instance selection mechanisms on \textit{mini-ImageNet} \cite{Vinyals2016}---a benchmark dataset often used for few-shot learning. We, further, use real-world data in the context of autonomous driving and foodservice. Specifically, we use the \textit{German Traffic Signs} dataset\footnote{\footnotesize\url{https://benchmark.ini.rub.de}} and the \textit{Food-101} dataset\footnote{\footnotesize\url{https://data.vision.ee.ethz.ch/cvl/datasets_extra/food-101/}}. We split each dataset into train, validation, and test classes. Hereby, we follow current literature that works with the mini-ImageNet dataset (train: 64 classes, validation: 16 classes, test: 20 classes) \cite{Ravi2017OptimizationAA} and apply a random split on the Food-101 set (65 / 16 / 20). We split the classes manually for the German Traffic Sign dataset (\eg, red signs are part of the train set and blue signs appear only in the validation and test sets resulting in 23 / 10 / 10).

\subsection{Metrics}
Following literature \cite{Pezeshkpour2020,Boney2017,Snell2017}, our evaluation is based on the accuracy metric. In the following, we elaborate its calculation: \textit{Accuracy} refers to the percentage of correct predictions $\Sigma$ given the total number of predictions $N$: $Accuracy = \frac{\Sigma}{N}$. Here, we distinguish between two types of accuracy, that is, method accuracy and model accuracy. Method accuracy refers to the accuracy when labeled instances (\ie, instances corrected by the human expert) are counted as correctly labeled, while model accuracy denotes the performance on the original query without considering the expert label in the evaluation. Thus, method accuracy represent the real-world HITL setting. Method accuracy can be regarded as the error correction rate and generally converges to an accuracy of 100\% as the human expert iteratively reviews the model predictions until all predictions are corrected. Note that we further validated our approach on the F1-Score, and came to the same conclusions as with the accuracy. For reasons of brevity, we, therefore, report our results on the accuracy metric as in \cite{Pezeshkpour2020,Boney2017,Snell2017}.

\subsection{Implementation details}

In the following, we describe our implementation. For the feature extractor, we draw upon a ResNet-12 \cite{he2016deep} including batch normalization and a ReLU activation. Overall, our model has several parallels to \cite{Oreshkin2018}. However, in our case, the batch normalization is not task-conditioned. For the ensemble-based techniques, we employ a model based on Monte-Carlo Dropout \cite{gal2017deep} with a dropout rate of $0.25$. We train our model in an episodic manner and sample for each task with $N$-way, $k$-shot and $b$-query, $N$ classes, $k$ images per class for the support set $\mathcal{S}$ and $b$ images per class for the query set $\mathcal{Q}$. We train our model in a 30-way, 10-shot, and 5-query setting informed by prior parameter studies. We validate and test the model within 5-way, 1-shot, and 15-query tasks as proposed in \cite{Snell2017} and draw upon this setting as the starting point for our HITL system. The computations are executed on a NVIDIA Tesla V100-SXM2 GPU with 32 GB RAM. {Code is available at \url{https://github.com/human-ai-research/HITL-few-shot-learning}.}

\section{Results}\label{sec:results}

{In the following, we present our results. First, we contrast the performance of our approach based on different instance selection mechanisms and, second, evaluate the cost-effectiveness of acquiring human expert knowledge in few shot image classification.}

\begin{figure}[H]
\captionsetup{position=top}
\centering
\subfloat[Popular image classification dataset]{
\begin{tabular}[c]{@{}ll@{}}
\textsc{MiniImagenet} & \textsc{MiniImagenet} \\ 
\includegraphics[width=0.4\linewidth]{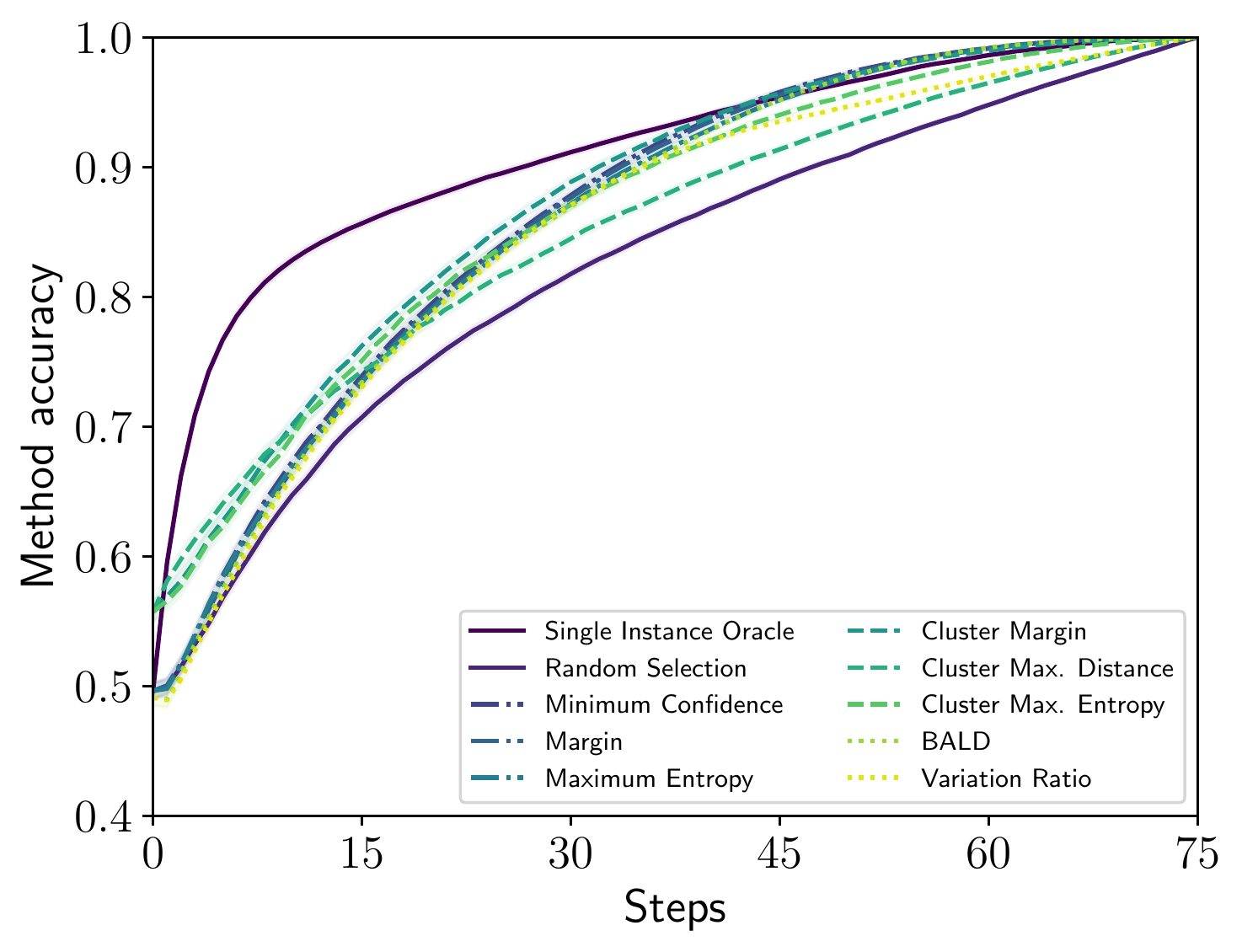} & \includegraphics[width=0.4\linewidth]{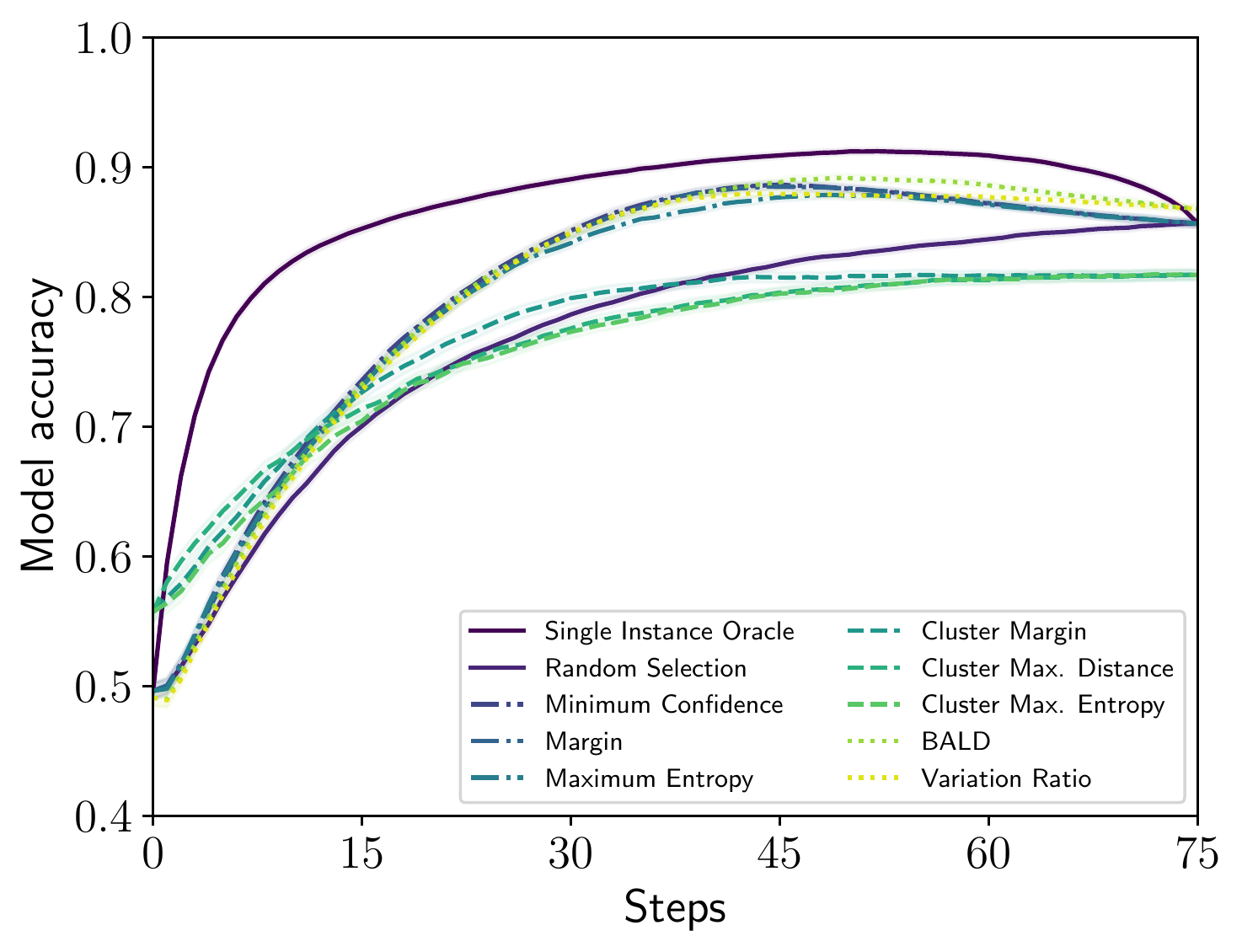}\\
\end{tabular}%
} \hfill\subfloat[Real-world industry datasets]{
\begin{tabular}[c]{@{}ll@{}}
\textsc{German Traffic Signs} & \textsc{German Traffic Signs} \\
\includegraphics[width=0.4\linewidth]{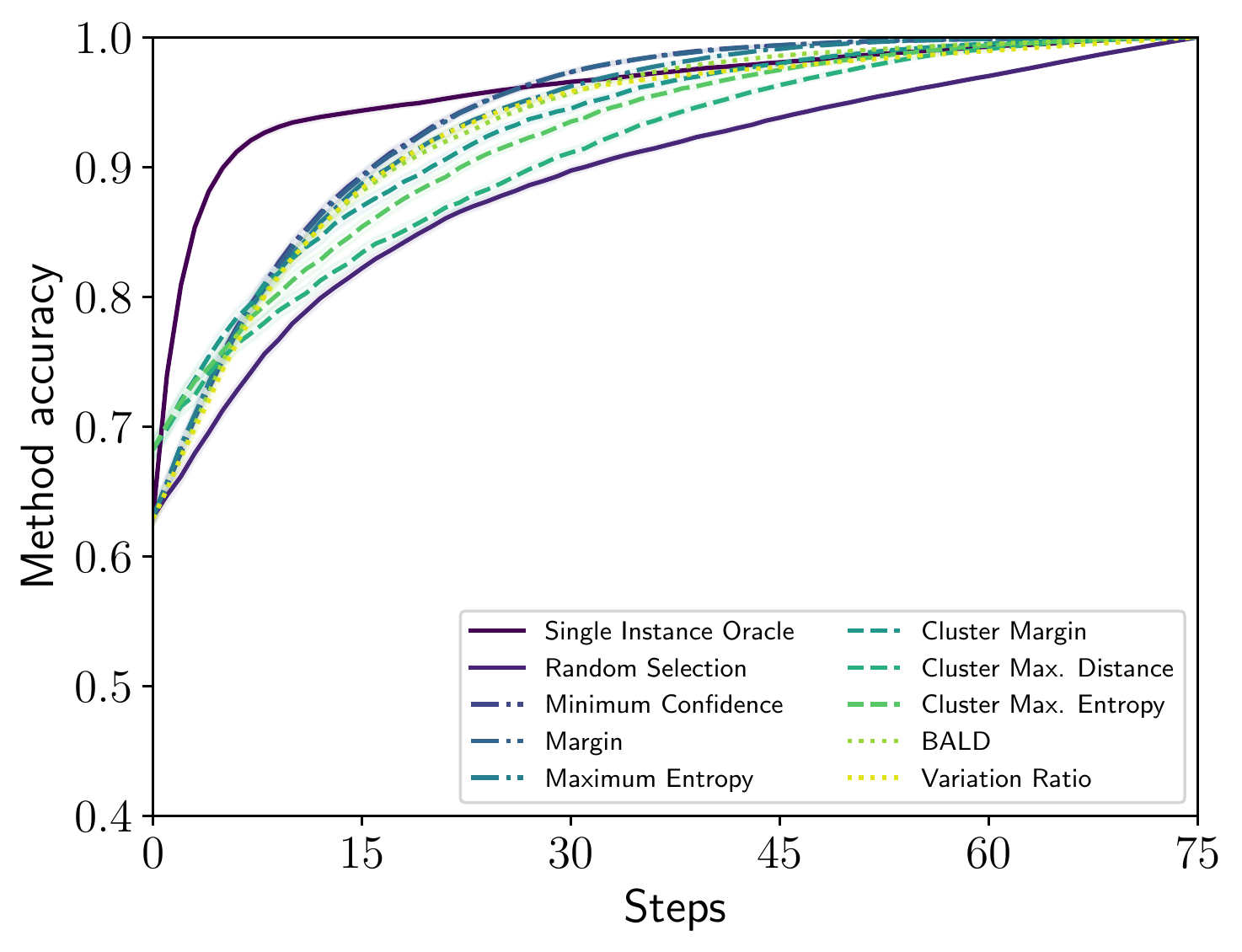} & \includegraphics[width=0.4\linewidth]{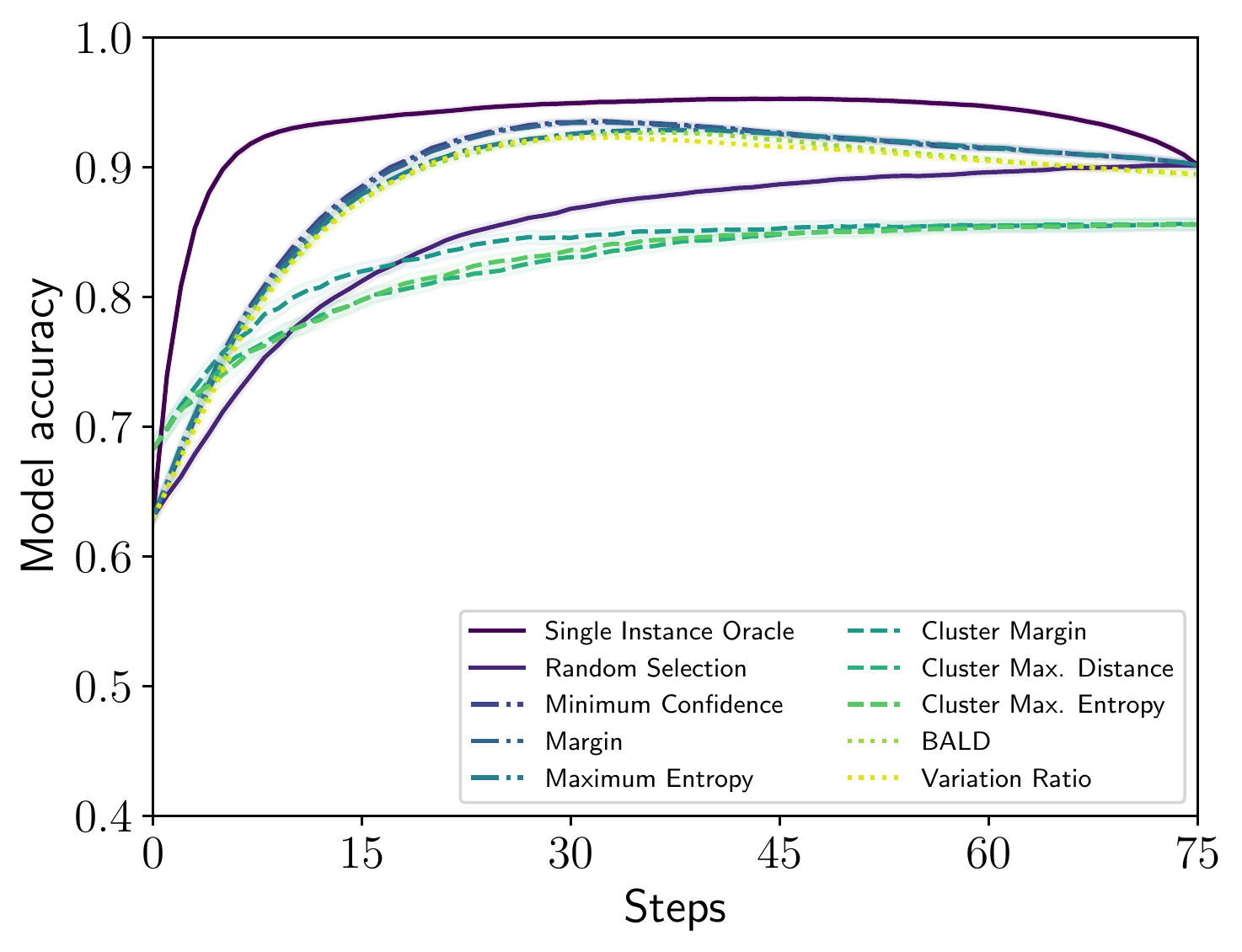}\\
\textsc{Food 101} & \textsc{Food 101} \\
\includegraphics[width=0.4\linewidth]{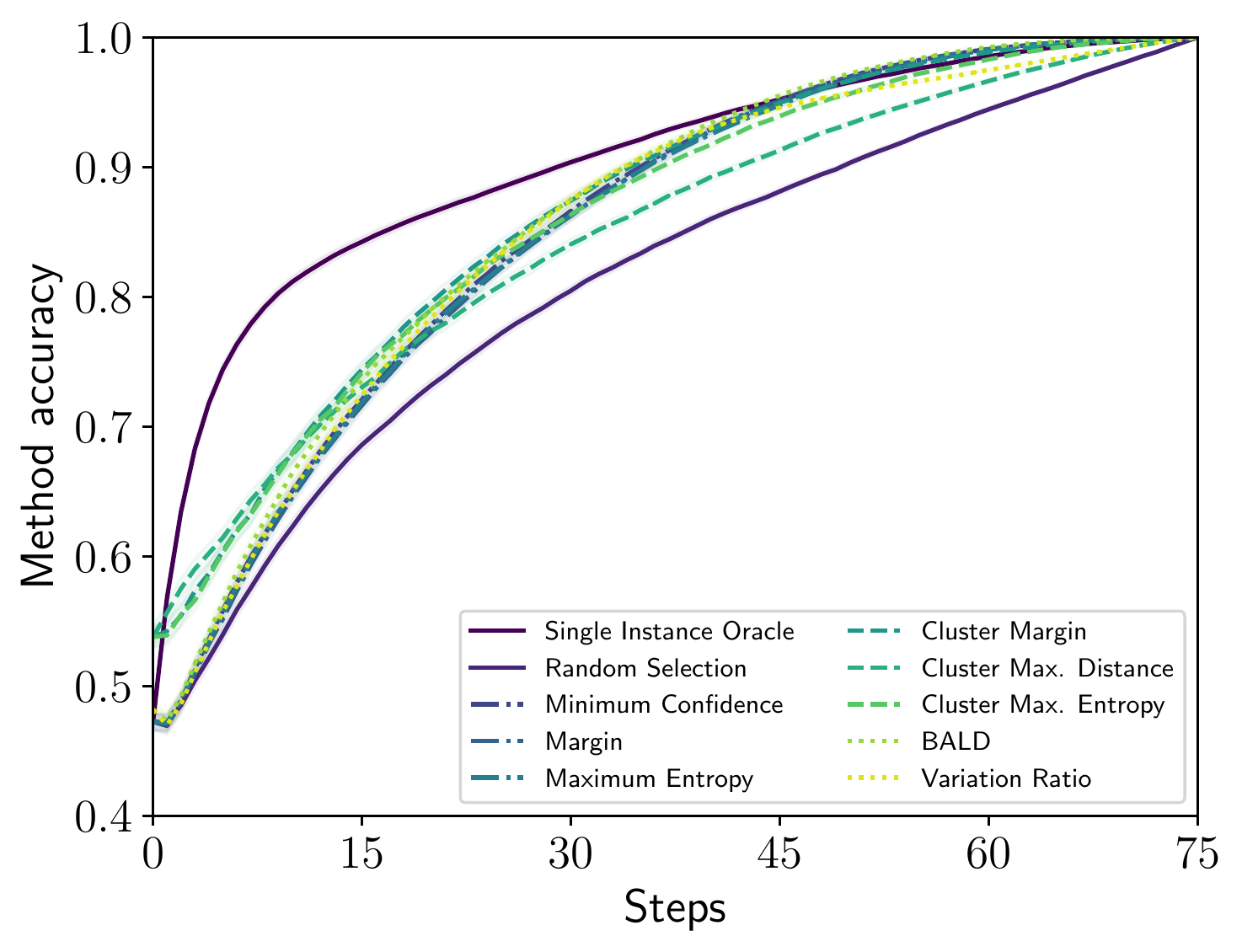} & \includegraphics[width=0.4\linewidth]{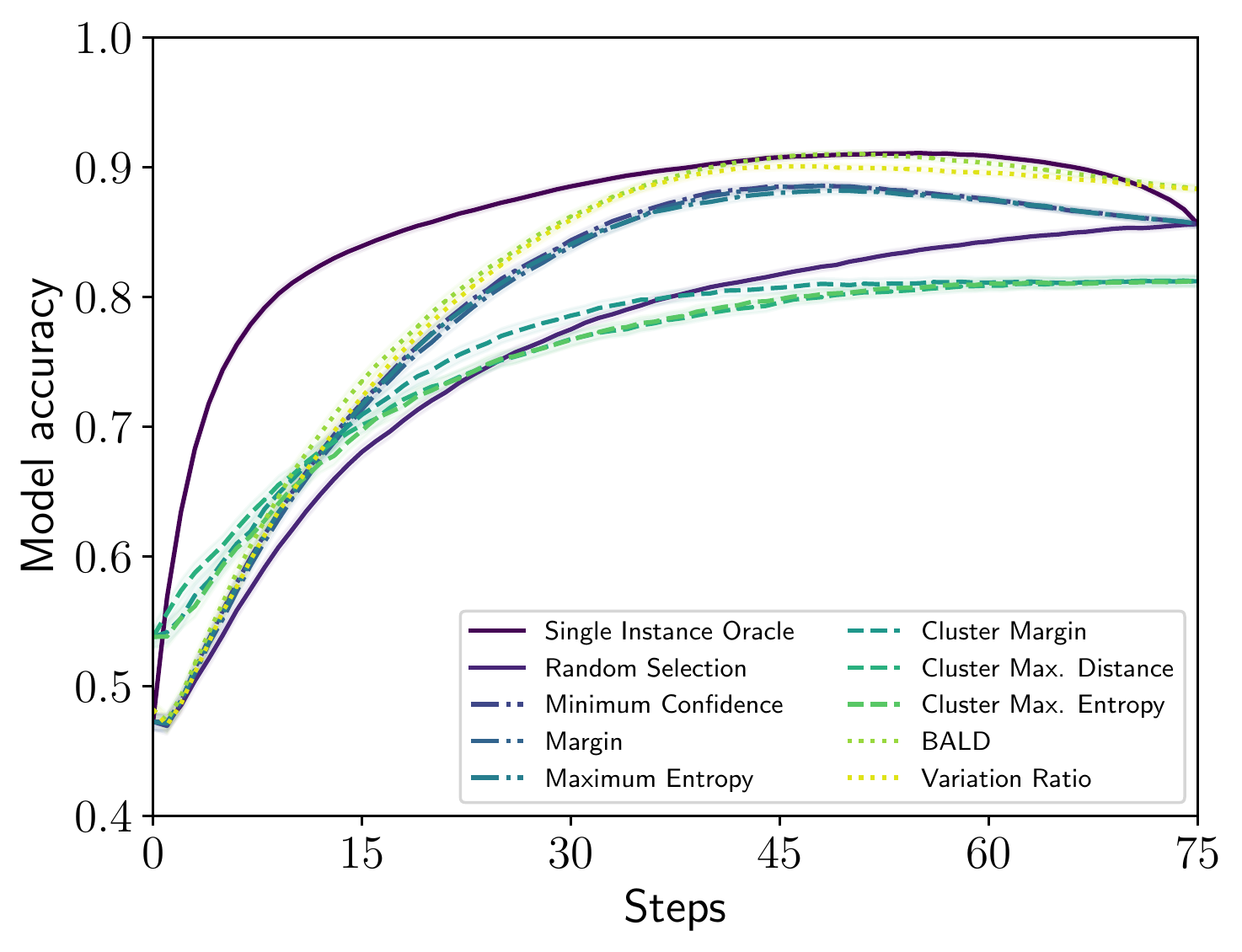} \\
\end{tabular}%
} \hfill
\caption{Experimental results of instance selection mechanisms in a HITL few-shot system for a 5-way 1-shot 15-query setting. The results are averaged over 1000 runs. Figures in the left column depict the experimental results evaluated on the method accuracy. 
Steps indicate the number of images labeled by the HITL system and the maximum number of steps denotes a human labeling budget of 100\%.}
\label{fig:results}
\end{figure}

\subsection{{Performance evaluation}}
We report the performances of different instance selection mechanisms in Table \ref{tab:accuracy} and Figure \ref{fig:results}, and focus on the required labeling costs. Overall, our evaluation suggests that confidence-based and ensemble-based instance selection mechanisms outperform random instance selection significantly (\eg, +3.0\% model accuracy for Minimum Confidence compared to random selection on Food-101 with -35.5\% of labeling effort). The ensemble-based instance selection mechanism BALD proposed for few-shot learning in this work achieves the best model performance for two of three datasets.

In the case of the Food-101 dataset, ensemble-based techniques achieve the model performance of the single instance oracle. Note that the model performance of ensemble-based approaches is not bounded by the single instance oracle, as the corresponding few-shot learning model includes a higher base performance due to the dropout of certain neurons of the network. This can also be observed in Table \ref{tab:model}, where the accuracy on Food-101 at 0\% budget is already higher for ensemble-based mechanisms than, for instance, confidence-based mechanisms. In this setting, our evaluation demonstrates that ensemble-based techniques are capable of outperforming confidence-based and cluster-based mechanisms. Further, we observe cluster-based approaches to surpass random selection and confidence-based instance selection for the first instances of the learning process due to increased initial performance. Here, our results are in line with previous findings on cluster-based few-shot learning \cite{Boney2017}. Thus, for low human labeling budgets, cluster-based mechanisms are beneficial. With an increasing human labeling budget, confidence-based and ensemble-based mechanisms should be preferred. Note that we further evaluated the HITL system on a separate hold-out query to further test the generalizability. We find consistent patterns for the instance selection mechanisms on the German traffic sign dataset. Here, all ensemble-based and confidence-based mechanisms outperform the single instance oracle given small labeling effort (\ie, after 20 steps). In contrast, on miniImagenet and Food-101, we observe less performance differences for the mechanisms on the hold-out set. Overall, we find that it is important for the HITL few-shot setting to have full access to all unlabeled instances during instance selection, which is in line with previous research (\eg, \cite{Pezeshkpour2020}).

\begin{table}[ht]
\centering
\scriptsize
\begin{tabular}{llcccc}
\toprule
Dataset & Mechanism & Accuracy at & Maximum & Budget at & Budget at \\
 &  & 5-Shot & accuracy & max. accuracy & 80\% accuracy \\
\midrule
\textsc{miniImagenet} & Single Instance Oracle &  86.9 ± 0.3 &  91.2 ± 0.3 &              68.4 &                   10.5 \\
                     & Random Selection &  73.8 ± 0.4 &  85.6 ± 0.3 &              98.7 &                   46.1 \\ 
                     & \textbf{Minimum Confidence} &  78.6 ± 0.5 &  \textbf{88.6 ± 0.3} &              59.2 &                   28.9 \\
                     & Margin &  78.2 ± 0.5 &  88.5 ± 0.3 &              57.9 &                   30.3 \\
                     & Maximum Entropy &  78.2 ± 0.5 &  87.8 ± 0.3 &              64.5 &                   30.3 \\ 
                     & Cluster Margin &  75.8 ± 0.6 &  81.7 ± 0.4 &              97.4 &                   40.8 \\
                     & Cluster Max. Distance &  74.0 ± 0.5 &  81.7 ± 0.4 &              96.1 &                   56.6 \\
                     & Cluster Max. Entropy &  73.6 ± 0.6 &  81.7 ± 0.4 &              94.7 &                   57.9 \\
                     & \textbf{BALD} &  78.3 ± 0.5 &  \textbf{89.2 ± 0.3} &              64.5 &                   28.9 \\
                     & Variation Ratio &  77.9 ± 0.5 &  88.0 ± 0.3 &              56.6 &                   30.3 \\
[2ex]
\textsc{Food 101} & Single Instance Oracle &  85.8 ± 0.4 &  91.1 ± 0.2 &              72.4 &                   11.8 \\
                     & Random Selection &  72.0 ± 0.5 &  85.6 ± 0.3 &              98.7 &                   50.0 \\
                     & \textbf{Minimum Confidence} &  77.2 ± 0.5 &  \textbf{88.6 ± 0.3} &              63.2 &                   31.6 \\
                     & Margin &  76.5 ± 0.5 &  88.6 ± 0.3 &              63.2 &                   31.6 \\
                     & Maximum Entropy &  77.1 ± 0.5 &  88.2 ± 0.3 &              64.5 &                   31.6 \\
                     & Cluster Margin &  74.3 ± 0.6 &  81.2 ± 0.4 &              93.4 &                   50.0 \\
                     & Cluster Max. Distance &  73.1 ± 0.6 &  81.2 ± 0.4 &              97.4 &                   64.5 \\
                     & Cluster Max. Entropy &  72.8 ± 0.6 &  81.2 ± 0.4 &              98.7 &                   60.5 \\
                     & \textbf{BALD} &  78.8 ± 0.5 &  \textbf{91.0 ± 0.3} &              65.8 &                   28.9 \\
                     & Variation Ratio &  78.0 ± 0.5 &  90.0 ± 0.3 &              61.8 &                   30.3 \\
[2ex]
\textsc{German Traffic Signs} & Single Instance Oracle &  94.2 ± 0.3 &  95.3 ± 0.2 &              61.8 &                    2.6 \\
                     & Random Selection &  83.8 ± 0.4 &  90.1 ± 0.3 &              98.7 &                   18.4 \\
                     & \textbf{Minimum Confidence} &  91.5 ± 0.3 &  \textbf{93.5 ± 0.2} &              42.1 &                   10.5 \\
                     & Margin &  91.2 ± 0.3 &  93.4 ± 0.3 &              42.1 &                   10.5 \\
                     & Maximum Entropy &  90.5 ± 0.4 &  92.8 ± 0.3 &              48.7 &                   10.5 \\
                     & Cluster Margin &  83.2 ± 0.6 &  85.6 ± 0.4 &              96.1 &                   14.5 \\
                     & Cluster Max. Distance &  81.1 ± 0.6 &  85.6 ± 0.4 &              90.8 &                   21.1 \\
                     & Cluster Max. Entropy &  81.5 ± 0.6 &  85.6 ± 0.4 &              97.4 &                   21.1 \\
                     & \textbf{BALD} &  90.1 ± 0.4 &  \textbf{92.7 ± 0.3} &              48.7 &                   10.5 \\
                     & Variation Ratio &  90.3 ± 0.4 &  92.3 ± 0.3 &              43.4 &                   11.8 \\
\bottomrule
\end{tabular}
\caption{Evaluation results reported as mean {model} accuracy $\pm$ $95\%$ confidence from 1000 runs and human labeling budget of the query set in \% in the 5-way 1-shot 15-query setting. Best practices for confidence-based and ensemble-based instance selection are highlighted in bold. Budget refers to the proportion of labeled data. {Confidence is calculated based on the t-test statistic.}} 
\label{tab:accuracy}
\end{table}

Among the confidence-based instance selection mechanisms, we observe slight differences in the model performance in Table \ref{tab:accuracy} and Figure \ref{fig:results}. Here, over all three datasets, the best performing mechanism is Minimum Confidence, followed by the Margin approach. Instance selection based on the maximum entropy exhibits a reduced performance. For the ensemble-based approach, we observe BALD to overcome the performance of Variation Ratio as reported in Table \ref{tab:accuracy}. In the case of cluster-based instance selection, the Cluster Margin approach outperforms the remaining cluster-based mechanisms over all three datasets. This observation aligns with previous results, in which Cluster Margin equally constitutes the best performing selection mechanism \cite{Boney2017}.

Note that we further evaluated instance selection mechanisms based on Gaussian processes due to their desirable statistical properties. Overall, we find that Gaussian process-based instance selection mechanisms fall short in performance compared to all instance selection mechanisms evaluated in this paper as their demand of labeled data is not fulfilled in the few-shot learning setting. We, therefore, do not include the results in this section. 
When evaluating our approach with the F1-score, we observe that the order of instance selection mechanisms remains identical and the curves are very similar to the ones for model accuracy. Overall, our experimental results demonstrate the relevance of acquiring human expert knowledge in few-shot learning in general and the importance of effective mechanisms to select instances for which human expert knowledge should be acquired. For the latter, our evaluation suggests that a HITL system for few-shot learning drawing on confidence-based or ensemble-based instance selection mechanisms outperforms the random baseline by a large margin and significantly reduces the gap to the single instance oracle.

\begin{table}[ht]
\centering
\scriptsize
\begin{tabular}{lrrrrrr}
\toprule
Model & \multicolumn{2}{l}{ \quad\textsc{miniImagenet}} & \multicolumn{2}{l}{ \quad\textsc{Food 101}} & \multicolumn{2}{l}{ \quad\textsc{German Traffic Signs}} \\
 & \quad0\% Budget & 100\% Budget &   \quad0\% Budget & 100\% Budget &  \quad0\% Budget & 100\% Budget \\
\midrule
Standard &         49.6 ± 0.6 &  85.6 ± 0.3 &  47.2 ± 0.7 &  85.6 ± 0.3 &           63.0 ± 0.6 &  \textbf{90.1 ± 0.3} \\
Ensemble &         49.1 ± 0.6 &  \textbf{86.8 ± 0.3} &  48.2 ± 0.6 &  \textbf{88.3 ± 0.3} &           62.9 ± 0.7 &  89.4 ± 0.3 \\
Cluster  &         \textbf{55.7 ± 0.8} &  81.7 ± 0.4 &  \textbf{53.8 ± 0.8} &  81.2 ± 0.4 &           \textbf{68.2 ± 0.9} &  85.6 ± 0.4 \\
\bottomrule
\end{tabular}
\caption{Evaluation results for different base models at 0\% and 100\% labeling budget reported as mean accuracy $\pm$ $95\%$ confidence over 1000 runs. Best performances per dataset are highlighted.  {Confidence intervals are calculated based on the t-test statistic.}}
\label{tab:model}
\end{table}

\begin{figure}
    \centering
    \includegraphics[width=0.4\linewidth]{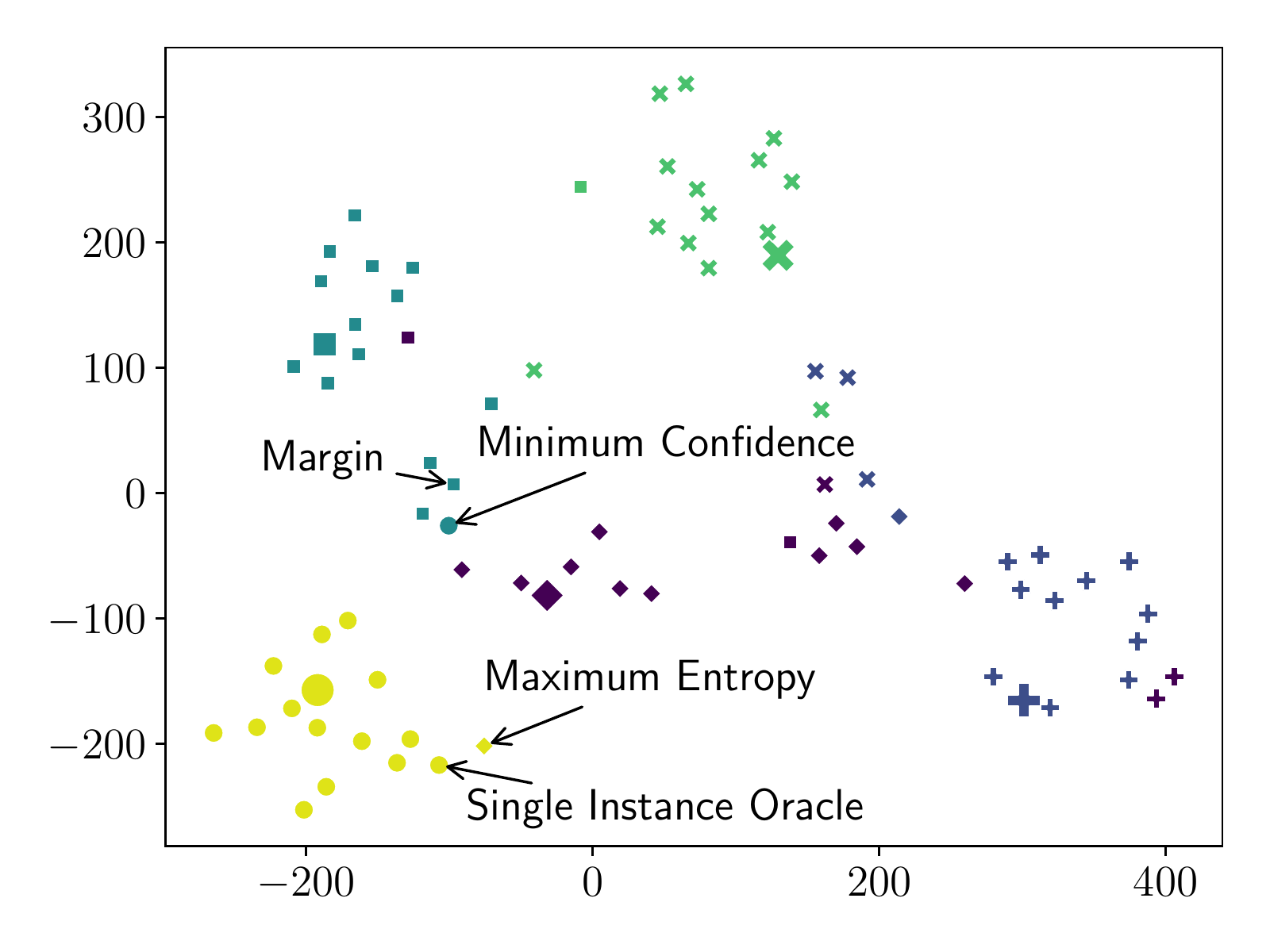}
    \includegraphics[width=0.4\linewidth]{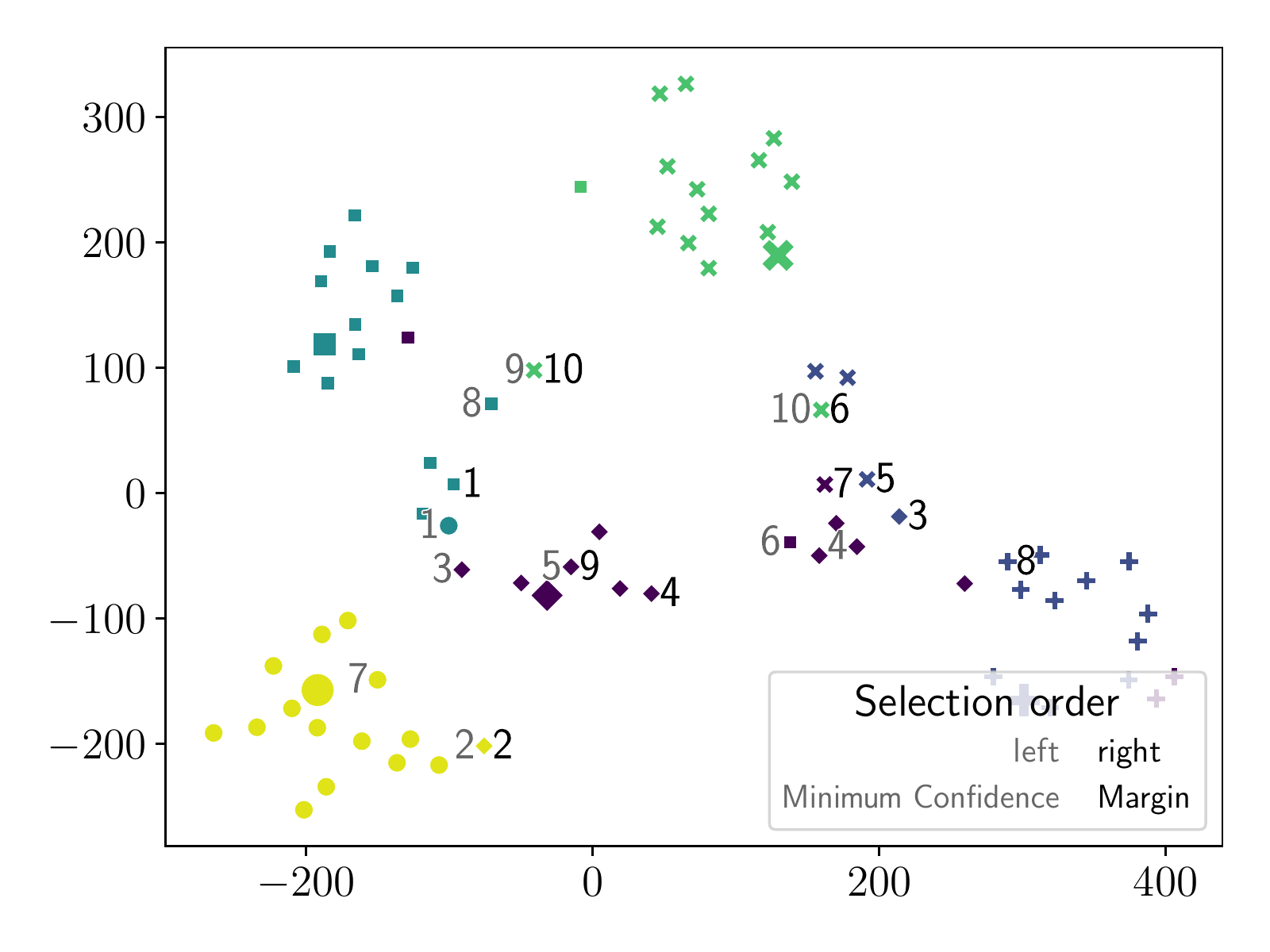}
    \caption{Visualization of instance selection mechanisms as part of an t-SNE representation \cite{maaten2008}. Color indicates the ground truth class, the symbol refers to the predicted class at 1-shot and large symbols denote the class prototype. In the left figure, arrows point to the first instance selected by each of the selection mechanisms. Numbers in the right figure indicate the first ten selections for the Margin mechanism and Minimum Confidence mechanism. Cluster-based and ensemble-based mechanisms are based on different models and are not shown for reasons of brevity.}
    \label{fig:tsne}
\end{figure}

In Figure \ref{fig:tsne}, we illustrate the meaningfulness of instance selection mechanisms in few-shot learning. Here, we depict a low-dimensional representation (\ie, t-SNE \cite{maaten2008}) of the images from novel classes. Each data point represents one image from the novel classes. The color indicates the specific ground truth class, the symbol refers to the predicted class after the 1-shot initialization, and the large data points represent the class prototype. We observe that two of the three instance selection mechanisms instantly select an instance with incorrect prediction without any knowledge about the ground truth. By handing these instances to the human expert, our model accelerates the performance.

\subsection{Cost-effectiveness of instance selection mechanisms in few-shot learning}
In this section, we study the cost-effectiveness of our HITL system for few-shot image classification. We assume constant costs per image and compare the accrued costs with the effectiveness in terms of the achieved model performance. Specifically, we provide two comparisons. That is, first, the costs at 80\% model accuracy across all instance selection mechanisms and, second, the costs at maximum model performance.

In Table \ref{tab:accuracy}, we observe that the labeling costs at 80\% model accuracy vary significantly among the different instance selection mechanisms. Again, the best-performing mechanisms Minimum Confidence and BALD require the least labeling effort. Compared to random instance selection, the two approaches incur labeling costs between -7.9\% and -21.1\%. Cluster-based mechanisms required significantly more labeled data to achieve 80\% model accuracy and, apart from Cluster Margin, reach the level of 80\% model accuracy later than random selection. This emphasizes that the acquisition of human expert knowledge is more cost-effective with efficient instance selection mechanisms. Second, at maximum model accuracy, the required labeling effort also varies significantly. As reported in Table \ref{tab:accuracy}, the best performing instance selection mechanisms Minimum Confidence and BALD achieve their maximum performance when less than 66\% of the data is labeled. For the German traffic sign dataset, Minimum Confidence achieves the maximum performance at 42.1\%, and BALD at 48.7\%. Our findings  indicate that our approach significantly accelerates the effectiveness of labeling. This is important for real-world applications, as our HITL few-shot system can lead to a significant reduction in labeling effort. 

\section{Discussion}\label{sec:discussion}

Our work has various implications for the development of HITL systems for few-shot learning. First, our findings emphasize the meaningfulness of combining HITL systems and few-shot learning when the model needs to be adapted to novel classes based on limited data. As these settings become increasingly important in real-world settings, our HITL few-shot system has desirable properties: For instance, for the German traffic sign dataset, the combination of human experts and the few-shot learning model achieves an increase of up to 48.4\% model performance which translates to a significantly improved method accuracy. Thus, our work demonstrates the meaningfulness of acquiring human expert knowledge. Second, we observe a sensitivity of the overall performance given different instance selection mechanisms. Notably, we developed ensemble-based instance selection mechanisms, which achieve the performance of the single instance oracle in parts of our experiments. Hence, we conclude that the successful interaction of human experts and few-shot learning models significantly benefits from meaningful instance selection mechanisms (\cf \cite{Zschech2021}). Third, our work provides a means to a cost-effective acquisition of human expert knowledge in few-shot learning. Overall, our approach caps labeling costs, which is desirable in real-world applications.

As with any other research, ours is not free of limitations. In the following, we name limitations and address directions for future research in the field of hybrid intelligence for few-shot learning. First, in our work, we follow literature and evaluate our few-shot learning model in the commonly used 5-way, 1-shot, 15-query setting \cite{Snell2017}. Other parameterizations are not discussed in this paper due to computational constraints. A second limitation of few-shot learning based on PNN, in general, is that the performance reduces with a higher number of classes. This constitutes a practical constraint to few-shot learning models and inherently for the HITL systems on top of the model, which practitioners should bear in mind. Traditional computer vision approaches should be consulted for high numbers of classes, while few-shot learning performs best when adapting a model to several novel classes. Third, our work draws upon the popular implementation of PNN. However, recent research developed additional few-shot learning models, which require further research in the context of HITL systems and hybrid intelligence.

This work includes several directions for future work. First, research on HITL systems and hybrid intelligence would benefit from demonstrating the cost-effectiveness of instance selection mechanisms within real-world use cases. Second, there is room for theoretical contributions on acquisition functions to close the gap between the oracle approach and recent instance selection mechanisms. Third, the rigorous design of an interface between few-shot learning systems and human experts may add to research in the field of hybrid intelligence and contribute to a broader understanding of the effective collaboration of humans and artificial intelligence.

\section{Conclusion}\label{sec:conclusion}

In this work, we propose a human-in-the-loop system for few-shot learning and provide a comprehensive evaluation of diverse instance selection mechanisms in a wide range of experiments from real-world image classification tasks. Our findings indicate that instance selection mechanisms significantly accelerate the overall performance and demonstrate the relevance of hybrid intelligence in few-shot learning. Furthermore, we show that instance selection mechanisms require considerably less data to achieve high model performances, which implies major cost savings in real-world applications. 

\section{Acknowledgments}

The authors acknowledge the financial support by the Federal Ministry for Economic Affairs and Climate Action of Germany in the project Smart Design and Construction (project number 01MK20016F).

\bibliographystyle{splncs04}
\bibliography{literature.bib}

\end{document}